\documentclass[manuscript,screen]{acmart}
\usepackage{tabularx}
\usepackage{booktabs}

\AtBeginDocument{%
  }

\setcopyright{acmlicensed}
\copyrightyear{2024}
\acmYear{2024}
\setcopyright{acmlicensed}\acmConference[SETN 2024]{13th Conference on Artificial Intelligence}{September 11--13, 2024}{Piraeus, Greece}
\acmBooktitle{13th Conference on Artificial Intelligence (SETN 2024), September 11--13, 2024, Piraeus, Greece}
\acmDOI{10.1145/3688671.3690604}
\acmISBN{979-8-4007-0982-1/24/09}

\begin{document}

\title{Prompt2Fashion: An automatically generated fashion dataset}

\author{Georgia Argyrou}
\email{argeorgiaa@gmail.com}
\authornotemark[1]
\affiliation{%
  \institution{National Technical University of Athens}
  \city{Athens}
  \country{Greece}
}

\author{Angeliki Dimitriou}
\email{angelikidim@ails.ece.ntua.gr}
\affiliation{%
  \institution{National Technical University of Athens}
  \city{Athens}
  \country{Greece}
}

\author{Maria Lymperaiou}
\email{marialymp@ails.ece.ntua.gr}
\affiliation{%
  \institution{National Technical University of Athens}
  \city{Athens}
  \country{Greece}
}

\author{Giorgos Filandrianos}
\email{geofila@ails.ece.ntua.gr}
\affiliation{%
  \institution{National Technical University of Athens}
  \city{Athens}
  \country{Greece}
}

\author{Giorgos Stamou}
\email{gstam@cs.ntua.gr}
\affiliation{%
  \institution{National Technical University of Athens}
  \city{Athens}
  \country{Greece}
}



\begin{abstract}
Despite the rapid evolution and increasing efficacy of language and vision generative models, there remains a lack of comprehensive datasets that bridge the gap between personalized fashion needs and AI-driven design, limiting the potential for truly inclusive and customized fashion solutions.
In this work, we leverage generative models to automatically construct a fashion image dataset tailored to various occasions, styles, and body types as instructed by users. We use different Large Language Models (LLMs) and prompting strategies to offer personalized outfits of high aesthetic quality, detail, and relevance to both expert and non-expert users' requirements, as demonstrated by qualitative analysis.
Up until now the evaluation of the generated outfits has been conducted by non-expert human subjects. Despite the provided fine-grained insights on the quality and relevance of generation, we extend the discussion on the importance of expert knowledge for the evaluation of artistic AI-generated datasets such as this one. 
Our dataset is publicly available on GitHub at \url{https://github.com/georgiarg/Prompt2Fashion}.
\end{abstract}



\begin{teaserfigure}
    \centering
    \includegraphics[width=\textwidth]{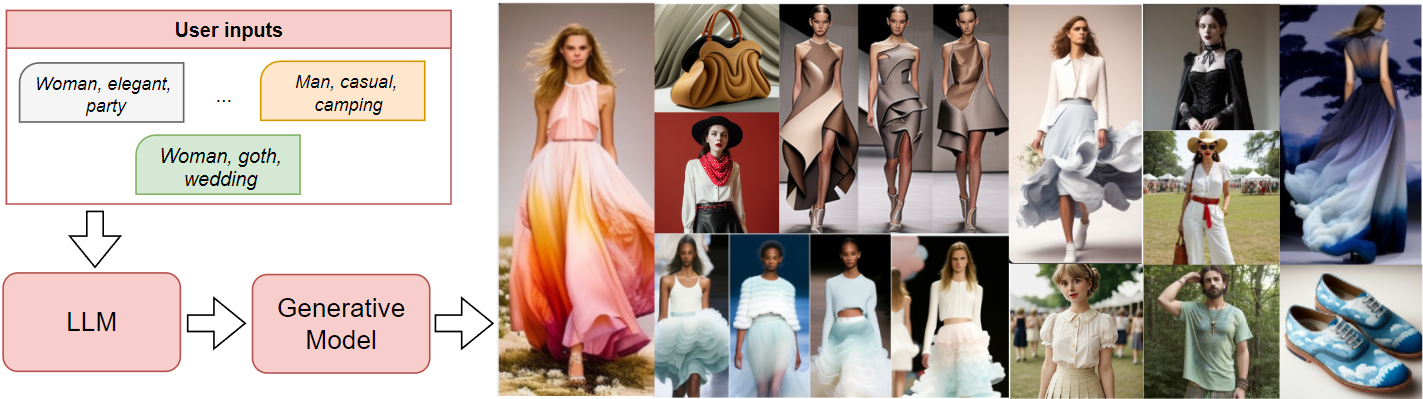}
    \caption{Automatic creation of fashion images using LLMs and diffusion models for generation.}
    \label{fig:teaser}
\end{teaserfigure}

\received{19 May, 2024}
\received[revised]{29 August, 2024}
\received[accepted]{15 July, 2024}

\maketitle

\section{Introduction}
The intersection of artificial intelligence (AI) and fashion is revolutionizing the industry by enhancing creativity, personalization, and efficiency. From designing garments to predicting trends, AI is becoming an indispensable tool for fashion designers, retailers, and marketers. However, the integration of AI into fashion faces significant challenges, particularly in evaluating AI-generated content, which often requires domain expertise to ensure relevance, style, and appeal. 

Leveraging these advancements, in this work, we created a dataset of fashion outfits that, rather than relying on existing annotated pictures, is entirely AI-generated. This approach allows us to produce a vast array of diverse images that meet various standards and personalization requirements, which would be costly and time-consuming to achieve through traditional methods. Moreover, we ensure the quality of these images by having them reviewed by human annotators, providing an added layer of validation.
The presented dataset encompasses a variety of characteristics, including gender, body type, occasions, and styles, along with their combinations. By leveraging the capabilities of Large Language Models (LLMs) followed by a Diffusion Model, we offer a scalable solution for generating fashion images. Our approach eliminates the need for human intervention in designing the final outfit or even the conditioning prompt to the Diffusion Model.

The scalability of production facilitated by LLMs and Diffusion Models ensures a diverse range of fashion images can be generated efficiently. The quality guarantees provided by the LLMs in language generation, as well as the Diffusion Models in image generation, are validated by human evaluators. This validation process reflects how potential consumers perceive these AI-generated outfits, ensuring that the content is not only technically proficient but also resonates with current fashion trends and consumer preferences. 

Given that AI is making significant inroads into creative fields, it is crucial that human oversight regulates generated content. After all, fashion image synthesis frameworks are ultimately designed for experts in the field, such as fashion designers. These AI-generated images are likely to serve as preliminary steps in the creative process rather than the final product. Consequently, it is essential for AI-generated fashion content to be evaluated by individuals with domain expertise. To this end, in this work, we not only offer an open, automatically generated dataset for creatives and engineers 
already reviewed by humans 
but also emphasize the importance of involving experts in the evaluation process, based not only on the aforementioned notions but also on experimental findings.

\section{Related work}
\paragraph{AI in fashion}
Very recent works leverage generative models for automatic fashion image generation, harnessing human perception for evaluation \cite{argyrou2024automaticgenerationfashionimages}. Nevertheless, the lack of fashion experts as evaluators limits the insights derived, since several fashion-related concepts and details cannot be addressed by non-experts. Other datasets  \cite{jia2020fashionpedia, 7780493, DBLP:journals/corr/abs-1710-07346, yu2024quality, huang2023first, morelli2022dresscode, zheng2019modanet, DBLP:journals/corr/abs-1906-05750} have been developed to support fashion image generation and various fashion-related tasks.

\paragraph{Evaluation of generative models}
In the realm of AI-generated fashion, evaluating the quality and relevance of the generated content is crucial, yet challenging. Common metrics such as Inception Score (IS) \cite{barratt2018noteinceptionscore} and Fréchet Inception Distance (FID) \cite{nunn2021compoundfrechetinceptiondistance} primarily measure image quality and diversity but fail to capture fashion-specific attributes like style consistency, trend relevance, and aesthetic appeal. Methods like the Visual Turing Test \cite{article}, which involves human evaluators determining the realism of images, and user study ratings, where participants assess attractiveness, trendiness, and wearability, provide general insights but lack the depth needed for detailed fashion critiques. These metrics are insufficient for fashion evaluation because they do not account for the subjective and nuanced nature of fashion, including cultural context, fabric draping, and fit. This inadequacy underscores the importance of fashion experts, who bring in-depth knowledge and a trained eye to evaluate AI-generated content. Fashion professionals can assess trend relevance, aesthetic and style consistency, cultural sensitivity, and technical aspects such as fabric representation and garment construction. Their expertise ensures that the generated designs meet industry standards and consumer expectations, bridging the gap between AI capabilities and the nuanced demands of the fashion industry.

\section{A generated fashion dataset}
In our study, we created a dataset of fashion images by using LLMs and a SoTA Generative Model to generate descriptions, which were then fed into a diffusion model, as presented in \cite{argyrou2024automaticgenerationfashionimages}. The input to our model consists of variable triplets. In our experiments, we use two types of triplets: "style, occasion, gender" and "style, occasion, type." The "type" variable includes both the body type and the gender of the wearer, such as "a small-framed delicate woman." This design allows us to observe how the model represents gender and adapts to different body types. These triplets are used to complete a custom prompt template, which varies depending on the prompting technique, to form a final prompt. This prompt is then fed into a LLM, and its output, which consists of the outfit description, serves as the input to a Generative Model, ultimately producing the generated image.

The LLMs used to produce the outfit descriptions are Mistral-7B \cite{jiang2023mistral} and Falcon-7B \cite{almazrouei2023falcon}, both of which are 7-billion-parameter language models. For the Generative Model, we used a Stable Diffusion model \footnote{\url{https://huggingface.co/emilianJR/epiCRealism}}.

In order to produce desirable results, we did not use neither training nor fine-tuning. We used different prompting techniques \cite{labrak2024zeroshot, DBLP:journals/corr/abs-2005-14165, chainofthought} to guide it and knowledge injection \cite{RAGmodel} to keep it up-to-date.

In order to construct the final triplets we used 2 simple types of people only differentiated by their gender ("man","woman"), 2 complex types of people which incorporated their body type in addition to their gender ("a small-framed, delicate woman", "a short, curvy man with a muscular build") and 10 occasions shown in table \ref{tab:occasions_styles}. We also used 5 styles shown in table \ref{tab:occasions_styles}. In our experiments we did all the possible combinations for these triplets resulting in 100 triplets with simple types and 100 with complex types. That lead to 200 triplets for each method, thus 1000 samples for every LLM, resulting in 2000 samples in total. Each image requires less than a minute to be produced with our model.

The final dataset consists of 2000 samples, each containing the LLM output, original triplet, as well as the diffusion model output image. Human evaluator ratings are available for several dataset instances and are to be enriched in the future.

\begin{table}[h!]
\centering
\begin{tabularx}{\textwidth}{X X}
\begin{tabular}{>{\raggedright\arraybackslash}p{6cm}}
\toprule
\textbf{Occasions} \\ \midrule
A music festival\\
A business meeting\\
A graduation\\
A bachelorette / bachelor party\\
A play / concert\\
A job interview\\
A work / office event\\
A tropical vacation\\
\bottomrule
\end{tabular}
&
\begin{tabular}{>{\raggedright\arraybackslash}p{6cm}}
\toprule
\textbf{Occasions} \\ \midrule

A cruise\\
A wedding\\ \midrule
\textbf{Styles} \\ \midrule
classic\\
gothic\\
bohemian\\
casual\\
sporty\\ \bottomrule
\end{tabular}
\end{tabularx}

\vskip 0.1in
\caption{Occasions and styles used to create the triplets}

\label{tab:occasions_styles}
\end{table}

\subsection{Qualitative results}

\begin{figure}[h]
\centering
\includegraphics[width=1\linewidth]{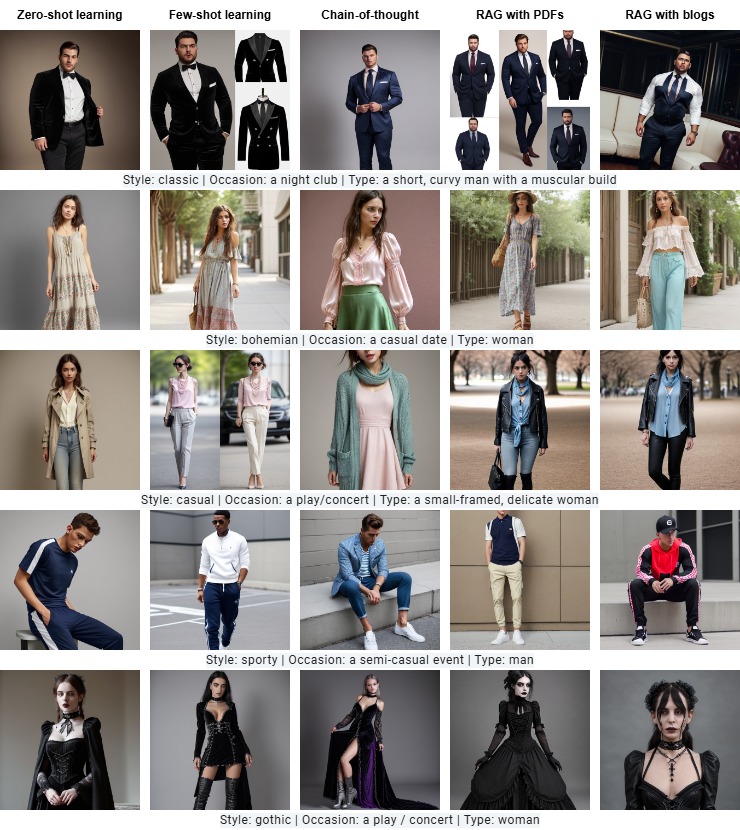}
\caption{Images produced by Stable Diffusion with descriptions generated with 5 different methods}
\label{fig:myres2}
\end{figure}

\begin{figure}[h!]
\centering
\includegraphics[width=1\linewidth]{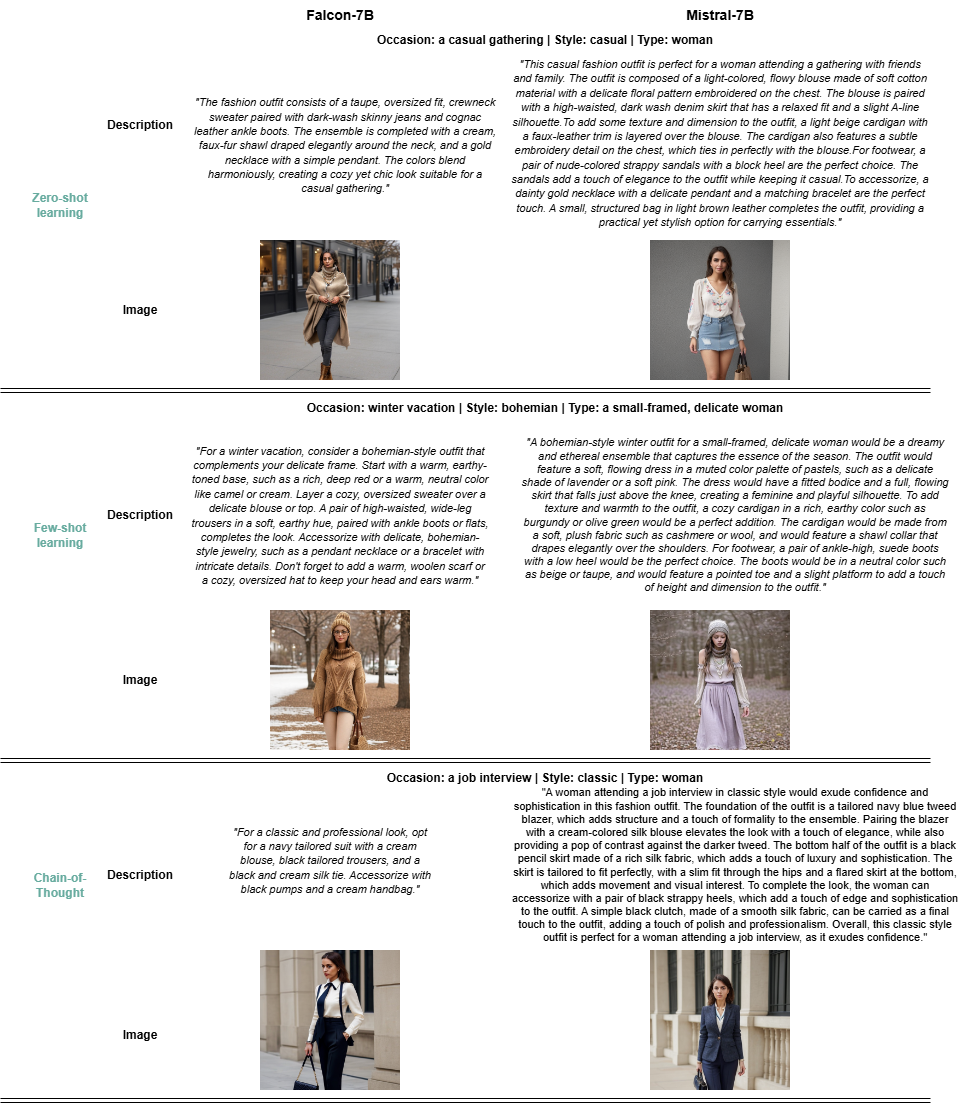}
\caption{Descriptions and images produced by Stable Diffusion with descriptions generated by 2 different LLMs for Zero-shot learning, Few-shot learning and Chain-of-Thought (left column for Falcon-7B model, right column for Mistral-7B model).}
\label{fig:comp1}
\end{figure}

\begin{figure}[h!]
\centering
\includegraphics[width=1\linewidth]{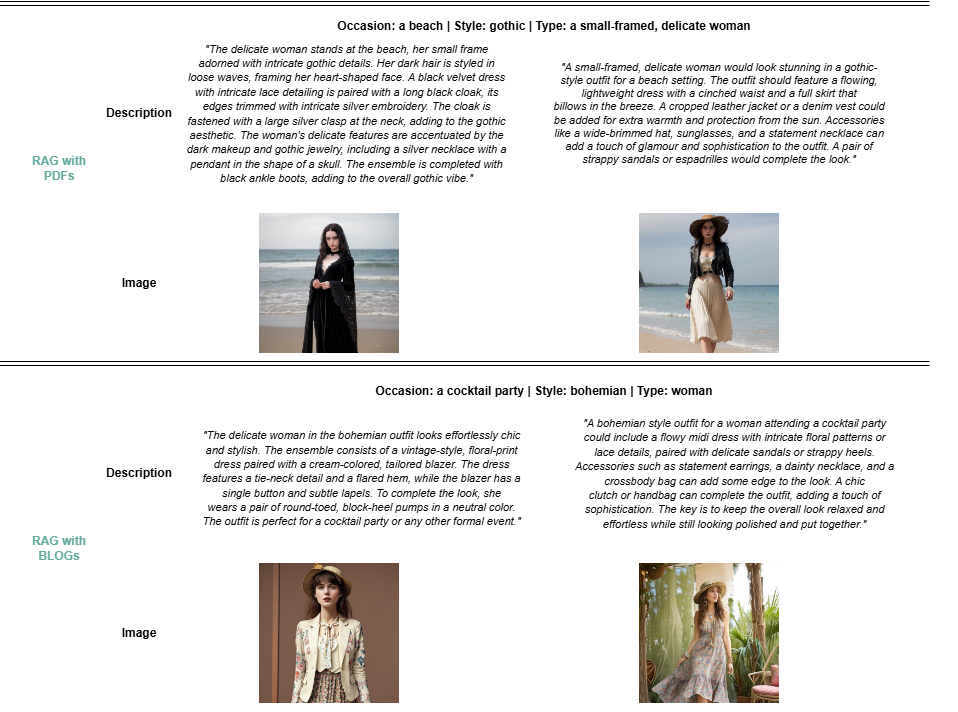}
\caption{Descriptions and images produced by Stable Diffusion with descriptions generated by 2 different LLMs for RAG with PDFs and with BLOGs as sources(left column for Falcon-7B model, right column for Mistral-7B model)}
\label{fig:comp2}
\end{figure}

In this section, we examine some samples of our dataset. In Figure \ref{fig:myres2}, images produced by Stable Diffusion by descriptions generated with 5 different methods are presented. In Figures \ref{fig:comp1} and \ref{fig:comp2}, the prompts are also presented in accordance with the images.

It is noted that many images subtly include the occasion in the background. For instance, a picnic is implied with a park setting, a winter vacation is suggested by a snowy landscape, and a festival is indicated by the presence of other people. These contextual clues in the background enrich the narrative and provide a more immersive experience for the viewer, helping to anchor the fashion description within a realistic scenario. In certain samples, multiple options are presented for the same triplet of variables. This variation arises because the generated descriptions might offer multiple choices for garments, colors, or textures. The diversity and contextual relevance of these samples make the dataset particularly useful for various applications. Engineers testing models can benefit from the diverse depictions of individuals and scenarios, which ensure robustness and adaptability in image generation models. For example, a small-framed delicate woman is not always depicted in the same way, providing a robust baseline for testing. Additionally, the inclusion of occasion-based backgrounds allows engineers to test how well their models maintain context relevance, crucial for applications like virtual try-ons or scenario-based advertising. Designers promoting diversity can use the dataset to showcase different body types, ethnicities, and fashion styles in various settings, aiding in the creation of more inclusive marketing materials and product designs. The contextual backgrounds can also help designers visualize their creations in real-world scenarios, which is particularly useful for seasonal collections or occasion-specific attire. Marketing and advertising professionals can leverage the contextual richness of the images to enhance storytelling and create more engaging advertisements by showing fashion items in realistic scenarios. Furthermore, the multiple options for garments, colors, and textures enable marketers to tailor campaigns to specific audience segments, broadening appeal and resonating better with diverse consumer groups. Academic researchers studying fashion trends and consumer behavior can use the dataset to analyze how different variables influence perception and preferences, while those focusing on artificial intelligence and machine learning can develop and test new algorithms for image generation and natural language processing. By providing a versatile and richly detailed dataset, we aim to support a wide range of applications and promote the development of more inclusive and context-aware models and designs.

\subsection{Can LLMs-as-Judges be used somewhere?}

To assess the overall quality of our results, we employed LLMs, specifically Mistral-7B, with the following prompt: \textit{"Respond with a score from 1 to 10 based on how well the response addresses the style, the occasion, and the role of the person wearing the outfit. It is advantageous if the response includes a variety of colors and textures."} However, LLMs not pretrained for this specific task often struggle to discern the fine-grained attributes of garments and assess the relevance of an outfit to a particular occasion or style, 
re-enforcing our belief that evaluation should be performed by humans.

\subsection{Non-expert evaluation}
Our qualitative findings and analyses have already been partially quantified in the previous work by \citet{argyrou2024automaticgenerationfashionimages}, which involved non-expert human annotators. The detailed rating process from that study is directly incorporated into this dataset.
In order to evaluate the generated results, non-experts were subjected to experiments to assess the descriptions and images produced. The evaluation results reveal insightful perceptions about the outfits presented in the generated images. The style of the outfits received a mean rating of 4.1 on a scale from 1 to 5, indicating that most participants found the style to align well with the intended design. The alignment of the outfits with the wearer's type received a high mean rating of 4.4, showing strong consensus on this match. Creativity, aesthetic appeal, and coherence were rated from moderate to very creative, and nearly all participants felt that the different garments and accessories matched well with each other, contributing to a cohesive and appealing overall look. The suitability of the outfits for various occasions indicated moderate agreement among participants. However, the textual descriptions of the outfits received higher scores for their suitability for the occasion, with a rating of almost 4.5. This suggests that the descriptions were more effective in conveying appropriate outfits for various occasions than the visual representations. Participants rated the outfit descriptions based on their suitability for the type, style, and occasion. The generated outfits performed well across these variables: the descriptions were highly suitable for the type, indicating a strong match with the described body types. Suitability for the style received strong ratings, showing alignment with the intended fashion styles. Colors described were considered appropriate for both the occasion and the type, reflecting thoughtful and appropriate color selection. The suitability of colors for the style was rated positively, demonstrating harmony between the color choices and the fashion styles described. Textures described were suitable for the occasion and well-matched with the type, suggesting they complemented various body types or fashion personas effectively. The suitability of textures for the style received strong ratings, affirming that the textures were seen as fitting and enhancing the described fashion styles. Overall, the high ratings across these criteria reflect a positive reception and suggest that the descriptions were effective in conveying the intended fashion concepts.

\subsection{Discussion: Expert evaluation}

Human evaluation is essential from an artistic perspective because fashion design is fundamentally a creative endeavor intricately tied to human expression. For this reason, it is crucial that the task be assessed by individuals who are not only potential users of these systems but also possess a deep understanding of artistic and aesthetic principles.
This approach enhances the relevance and practicality of the results in real-world applications, making them more useful and inspirational for those engaged in the fashion industry. 

Additionally, the wide variance in ratings from non-experts highlights the subjective nature of their evaluations and suggests a potential lack of deep understanding of fashion nuances. In contrast, expert evaluations offer a more complex and consistent perspective. Experts, with their extensive knowledge and experience, are better equipped to provide detailed assessments of design elements that non-experts might overlook. Their feedback is not only grounded in a thorough understanding of current trends and cultural relevance but also considers practical applications, ensuring that the designs are both innovative and marketable. 



The contrast between non-expert and expert evaluations becomes clear when examining their responses to the quality of generated images. A significant 78\% of non-experts did not observe any abnormalities or inconsistencies, indicating that, from a broad perspective, the images are seen as effective and visually appealing. Among the 22\% who did notice inconsistencies, most considered these issues minor, not enough to detract from the design's potential as a source of inspiration. However, this general approval from non-experts also highlights the limitations of relying solely on their evaluations, as they may lack the refined eye needed to detect subtleties that could affect the functionality, marketability, or artistic integrity of a design. In contrast, experts, with their deep understanding of design principles and industry standards, can identify and assess these nuances and inconsistencies, ensuring that they do not compromise the overall appeal of the design, its adherence to current trends or cultural relevance.


Incorporating expert evaluations into the design process makes it more rigorous and aligned with professional fashion standards. Experts can validate or reject the designs identified by non-experts, offering a more precise and reliable assessment of their artistic and commercial viability. This validation enhances the credibility of the evaluation and ensures that the final outputs meet high artistic standards and are practical for real-world applications. Additionally, expert feedback provides critical insights for refining the designs, offering specific recommendations that elevate the design from a preliminary concept to a polished, market-ready product.

In addition, it is noteworthy that non-expert participants rated the style of the outfits highly, with a mean rating of 4.1 out of 5. This high rating suggests a strong alignment with the intended design and indicates that, on a surface level, the generated outfits resonate well with the audience. However, this raises an important consideration: while the overall style may be appealing, the deeper elements that constitute and define a particular style are often not well understood by non-experts. Even though participants were provided with example outfits from each style as a ground truth, their understanding of the nuances and complexities involved in accurately evaluating these styles might be limited. Experts can critically assess whether the generated designs authentically capture the essence of the intended style or if they merely approximate it on a superficial level.

Moreover, there was a strong consensus among non-expert participants regarding how well the outfits matched the wearer's body type, with a mean rating of 4.4 out of 5. This consensus may be influenced by a general lack of knowledge about what truly suits different body types, as well as a tendency toward aesthetic biases, such as the "pretty bias," where visually appealing designs are automatically assumed to be well-suited to any wearer. Fashion experts, who have a deep understanding of body types and the principles of garment fitting, can provide a more accurate assessment of whether the designs genuinely complement the wearer’s body type. They are well-versed in how different cuts, fabrics, and styles interact with various body shapes to enhance or detract from the overall appearance.

When it comes to evaluating outfit descriptions, participants rated the comprehensibility and cohesion of the generated descriptions highly, with both aspects receiving an average score of approximately 4.3 out of 5. This positive feedback indicates that, overall, the descriptions were perceived as clear and well-structured by the participants. Notably, those who identified as proficient or native/fluent in English provided even higher scores compared to the rest. This distinction highlights the critical role of language proficiency in the evaluation process, suggesting that the descriptions were generally well-crafted and accessible to those with strong language skills. However, it's important to consider that fashion-related terminology and nuances might not be fully understood or appreciated by all participants, particularly those without specialized knowledge in fashion. While the descriptions may be comprehensible in a general sense, the use of specific fashion terms—such as the names of fabrics, cuts, or styles—may not be fully grasped by non-experts. This could lead to an overestimation of the descriptions' effectiveness by participants who are not familiar with the subtleties of fashion language. This is where expert evaluation becomes essential. Fashion experts, who are well-versed in industry-specific terminology and the nuances of style descriptions, can provide a more accurate and detailed assessment of the descriptions' quality. They can evaluate whether the terminology used is not only accurate but also appropriately conveys the intended style and design elements. Experts can also identify if the descriptions are effectively communicating the unique aspects of the outfits, ensuring that they resonate with both professionals and enthusiasts in the fashion industry.

By examining several observed patterns of the raters combined with the nature of the generated content, we believe that our dataset should be evaluated by fashion experts in the future and enriched with their scores and detailed ratings to enhance its utility and precision.

\section{Conclusion}

This study presents a novel approach to generating a comprehensive fashion dataset consisting of 2,000 images and descriptions. By leveraging LLMs and Diffusion Models, we have created a diverse collection of fashion content tailored to various occasions, styles, and body types. Our evaluations indicate that the generated images and descriptions are both relevant and aesthetically pleasing. While non-expert feedback confirms their appeal, expert evaluation is essential for ensuring high standards of fashion quality and marketability. 
Therefore, future work will focus on incorporating expert ratings to further refine the dataset and enhance its utility. Overall, this dataset represents a significant advancement in AI-driven fashion design, providing a valuable resource for further research and applications in the industry.

\begin{acks}
Maria Lymperaiou was supported by the Hellenic Foundation
for Research and Innovation (HFRI) under the 3rd Call for HFRI PhD Fellowships
(Fellowship Number 5537). 
Angeliki Dimitriou was supported by the Hellenic Foundation
for Research and Innovation (HFRI) under the 5th Call for HFRI PhD Fellowships
(Fellowship Number 19268). 
We would also like to thank the human annotators for their invaluable feedback.
\end{acks}

\bibliographystyle{ACM-Reference-Format}
\bibliography{sample-base}


\end{document}